# A spatiotemporal fused network considering electrode spatial topology and time-window transition for MDD detection

Chen-Yang Xu[a], Han-Guang Wang[a], Lan Zhang[a], Yong-Hui Zhang[b], Hui-Rang Hou[a], and Qing-Hao Meng[a*]

[a] *School of Electrical and Information Engineering, Tianjin University, Tianjin 300072, China*

[b] *Department of Psychiatry, Tianjin Anding Hospital, Tianjin 300021, China*

[*] Corresponding Author: Qing-Hao Meng, Email: qh_meng@tju.edu.cn

*Abstract*—**Recently, researchers have begun to experiment with deep learning-based methods for detecting major depressive disorder (MDD) using electroencephalogram (EEG) signals in search of a more objective means of diagnosis. However, existing spatiotemporal feature extraction methods only consider the functional correlation between multiple electrodes and temporal correlation of EEG signals, ignoring the spatial position connection information between electrodes and the continuity between time windows, which reduces the model's feature extraction capabilities. To address this issue, a Spatiotemporal fused network for MDD detection with Electrode spatial Topology and adjacent TIME-window transition information (SET-TIME) is proposed in this study. SET-TIME is composed by a common feature extractor, a secondary time-correlation feature extractor, and a domain adaptation (DA) module, in which the former extractor is used to obtain the temporal and spatial features, and the latter extractor can mine the correlation between multiple time windows, and the DA module is adopted to enhance cross-subject detection capability. The experimental results of 10-fold cross-validation show that the proposed SET-TIME method outperforms the state-of-the-art (SOTA) method by achieving MDD detection accuracies of 92.00% and 94.00% on the public datasets PRED+CT and MODMA, respectively. Ablation experiments demonstrate the effectiveness of the multiple modules in SET-TIME, which assist in MDD detection by exploring the intrinsic spatiotemporal information of EEG signals.**

***Keywords—MDD detection, EEG, Spatiotemporal fused Network.***

# I. INTRODUCTION

Major depressive disorder (MDD) is a common and severe mental disorder characterized by persistent feelings of sadness and a loss of interest or pleasure in activities that were previously rewarding or enjoyable. Existing methods usually use questionnaire surveys to diagnose MDD, including the Beck Depression Inventory [1], Patient Health Questionnaire [2], Diagnostic and Statistical Manual of Mental Disorders [3] and Hamilton Depression Rating Scale [4]. However, questionnaire-based diagnoses could bring subjective biases and may be denied by subjects [5]. With the maturity of electroencephalogram (EEG) technology, combining machine learning or deep learning to objectively diagnose MDD using EEG has become an effective alternative method[6][7][8].

Multiple manual features obtained from prior knowledge are required when integrating EEG data with traditional machine learning algorithms, such as support vector machine, random forest, K-nearest neighbor, and extreme gradient boosting. Ravan et al. [9] extracted symbolic transfer entropy from EEG data, and employed a feature selection strategy aimed at maximizing correlation with the target class while minimizing redundancy. The selected features were subsequently combined with classifiers to distinguish individuals with MDD from those with bipolar disorder. Li et al. [10] extracted power spectral density, fuzzy entropy, and phase lag index from EEG signals. After ranking the features, they were fed into classifiers such as support vector machine and random forest to obtain depression detection results. However, manually extracted features are highly dependent on prior knowledge. For example, for the multi-channel time-series of EEG data, its multiple channels and time periods must be carefully analyzed by domain experts before meaningful features can be extracted.

With the rise of deep learning, its powerful feature extraction capabilities [11] and end-to-end characteristics provide new ideas for MDD detection. Some deep learning methods are utilized to further extract the spatial and/or temporal features in depression detection tasks. Shen et al. [12] employed a hybrid approach combining multi-scale convolutional neural networks (CNN) and long short-term memory (LSTM) to capture both localized features and temporal correlations within each hemisphere of the brain. To enhance the effectiveness of spatial feature extraction, a cross-attention mechanism was applied to facilitate information interaction between hemispheric features. Xu et al. [13] utilized a graph convolutional network (GCN) to extract spatial features of the EEG signals and used a CNN to further weight the feature maps. Choudhary et al. [14] proposed a CNN-LSTM-based framework incorporating two distinct blocks, each designed to separately extract spatial and temporal features. The extracted features were subsequently concatenated to improve the performance of depression classification.

However, all these methods ignore core features of EEG data: its inherent spatiotemporal connections. Specifically, how EEG electrodes are positioned relative to each other, and how to establish a connection between two adjacent time windows while considering their transition period (i.e., the time span between the end of the previous time window and the beginning of the next time window). These connections are crucial for accurate EEG analysis [15] [16], and research in this critical area is still insufficient. In this study, we aim to enhance spatial information representation through electrode physical location information and enhance temporal dependency through time window correlation. Under this motivation, a Spatiotemporal fused Network for MDD detection with Electrode spatial Topology and adjacent TIME-window transition information, i.e., SET-TIME, is proposed. SET-TIME consists of a common feature extractor, a secondary time-correlation feature extractor, and a domain adaptation (DA) module.

The common feature extractor first employs multi-scale depth-wise convolution operations to extract multi-scale temporal features within each time window without disrupting the EEG electrode channels, thereby extracting high-semantic features while preserving high-resolution features. Then, the time-window transitional module is put forward to extract features from the interaction of two neighboring time windows. Third, the electrode spatial topology module containing the information about the spatial location and physical connection of the electrodes is proposed to mine spatial features.

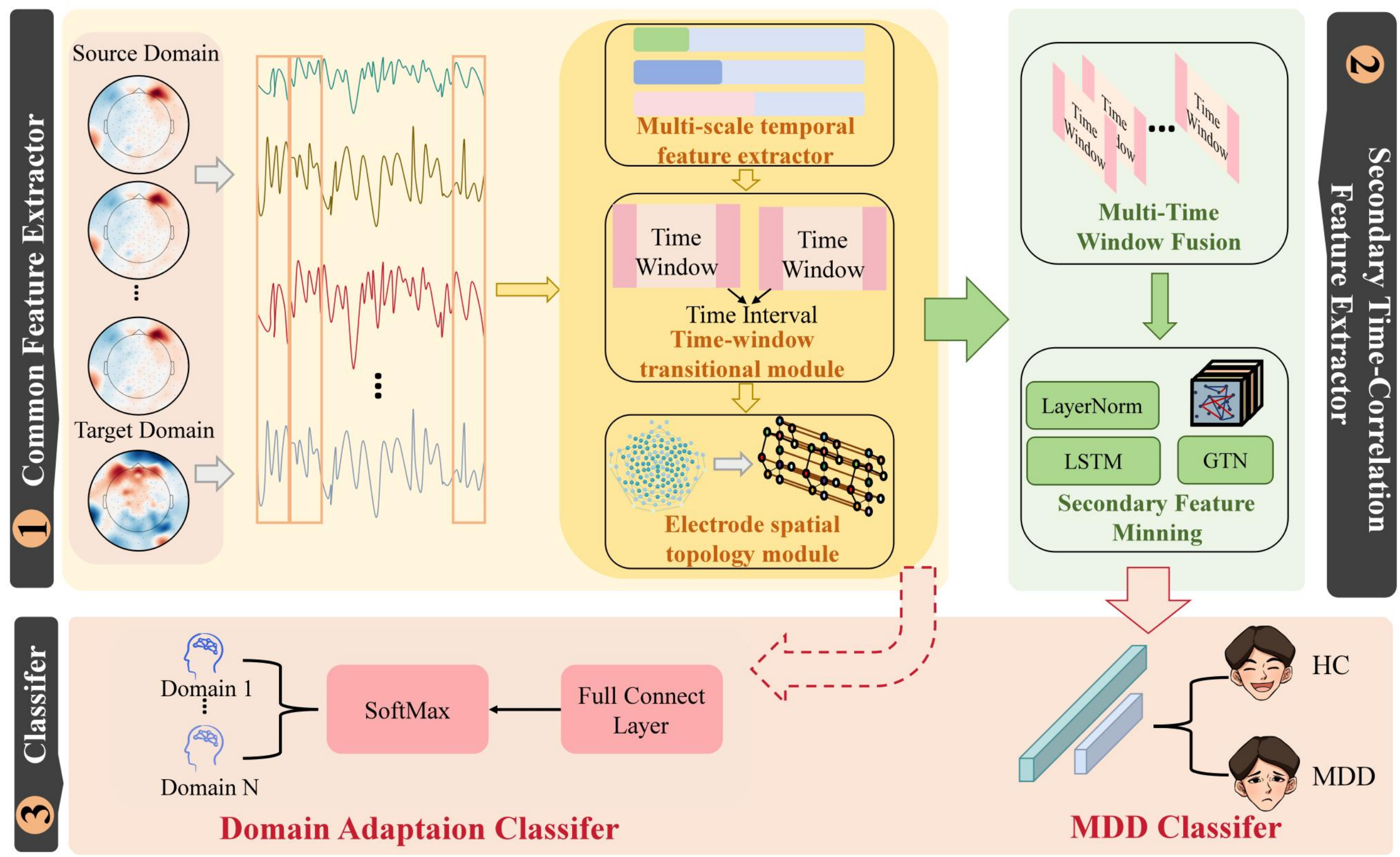

Figure 1. An overview of the proposed SET-TIME framework.

The secondary time-correlation feature extractor is proposed to mine deep temporal correlation information among multiple time windows, which can further enhance the temporal feature extraction capability and improve the discriminative capability of the downstream classifiers. The secondary time correlation extraction module provides temporal features with high resolution and semantic information. Meanwhile, the long-range temporal dependencies of multiple time windows can be further exploited. Finally, a domain adversarial learner-based domain adaptation (DA) module is adopted in this study to improve cross-subject MDD detection accuracy.

The main contributions of this paper include the following points:

(1) The time-window transitional features from the interaction of two neighboring time windows are extracted for the first time for MDD detection, which can enhance the long-range dependency without disrupting time-series order of EEG signals.
(2) An electrode spatial topology module that contains the spatial location and physical connection information is proposed to mine spatial features of EEG signals. This module combines trainable parameters with electrode positions to adaptively learn the optimal spatial connection topology between electrodes.
(3) The secondary time-correlation feature extractor is proposed to mine deep temporal correlation information among multiple time windows, which can further enhance the temporal feature extraction capability and improve the discriminative capability of the downstream classifiers.
(4) We conducted extensive experiments on two publicly available MDD datasets, and the results show that our approach outperforms the state-of-the-art (SOTA) approaches. Ablation experiments demonstrate the effectiveness of our method. We have also used t-SNE plots demonstrating the feature distribution of the model.

## II. METHODOLOGY

This chapter presents the SET-TIME model architecture, which comprises three core components: (1) Common feature extractor, which contains a multi-scale time feature extraction, a time-window transitional module and an electrode spatial topology module; (2) Secondary time-correlation feature extractor; (3) Domain adaptation Module.

The data input of the proposed framework is $\boldsymbol{X} \in \mathbb{R}^{B\times T\times V\times len}$, where $B$ is the batch size, $T$ indicates the number of time windows divided from the original EEG data, $V$ represents the number of EEG channels corresponding to EEG electrodes, and $len$ denotes the data length of one time window after division.

### *A. Common Feature Extractor*

#### *a) Multi-Scale Time Feature Extractor*

We use a common feature extractor for mining EEG features in the source and target domains. First, we extract the EEG features within an electrode channel using a multi-scale-depth-wise convolution operation. This ensures that the extracted features contain only the information of a single electrode and do not contain the features of other electrode channels, while the use of multi-scale convolution operation preserves the local features with low semantics and mines the long-range dependency features with high semantics. The depth-wise 1D-CNN kernels on a single scale are summarized as $\boldsymbol{CK} \in \mathbb{R}^{(V\times kd)\times 1\times ks}$, where the number of input channels is $V$, $ks$ and $kd$ indicate the size and the output dimension of the kernel, respectively. For the $kd$-dimension kernels of each input channel, the results are not added together but concatenated instead. That means at one EEG channel, the feature dimension is $l^{'} * kd$, where $l^{'}$ is the length of the signal following convolution with a kernel of size $ks$. Furthermore, we use a multi-scale convolutional operation to extract EEG features with different resolutions, enriching the feature space in the downstream network so that the model contains both high-semantic and high-resolution features. The detailed information of different depth-wise convolutional layers is shown in Table I. The final feature from this sector is represented as $\boldsymbol{f}_{depth} \in \mathbb{R}^{B\times T\times V\times\left(kd\times\sum_{i=1}^{3} fl_i\right)}$, where $fl_i(i=1,2,3)$ represents the feature length after processing with three different scale kernels.

TABLE I. Detailed information of different layers of depth-wise convolution using 1D-CNN.

| Layer number | Input channels | Kernel size | Output channels | Stride | Padding |
|---|---|---|---|---|---|
| | | 64 | | 8 | |
| 1 | $V$ | 32 | $V\times kd$ | 4 | valid |
| | | 16 | | 2 | |
| Batch Normalization, Activation: Relu, Maxpooling: 2 | | | | | |
| | | 16 | | | |
| 2 | $V\times kd$ | 8 | $V\times kd$ | 2 | valid |
| | | 4 | | | |
| Batch Normalization, Activation: Relu, Maxpooling: 4 | | | | | |
| | | 4 | | | |
| 3 | $V\times kd$ | 4 | $V\times kd$ | 1 | same |
| | | 4 | | | |
| | | 8 | | | |
| 4 | $V\times kd$ | 8 | $V\times kd$ | 1 | same |
| | | 8 | | | |
| Batch Normalization, Activation: Relu, Maxpooling: 4 | | | | | |

### b) *Time-Window Transitional Module*

The time-window transitional module uses an auto-encoder to extract features for the start and the end of each time window, and then digs deeper into the time window correlations to establish long-range dependencies. This method can avoid data redundancy caused by existing sliding window operations and the need to manually adjust the overlapping size [31]. Finally, all temporal correlation features are concatenated, ensuring that the temporal order is maintained. The time-window transitional module is summarized in Fig. 2, which is further divided into two parts. The first part generates the embeddings of the separate first start and last end slices by using two different linear layers. The outputs are separately $\boldsymbol{f}_{start}$ and $\boldsymbol{f}_{end} \in \mathbb{R}^{B\times 1\times V\times(2\times ts)}$, where *ts* indicates the time length of the start or end segment of each time window. The second part employs a dual-path autoencoder to embed the temporal features at the beginning and end of the sequence, thereby preserving the sequential order of the time windows. The last dimensions of the time window transitional module outputs are concatenated together to get the embeddings $\boldsymbol{f}^{i}_{interval} \in \mathbb{R}^{B\times l\times V\times(2^{*}ts)}(i = 1,2,\ldots,T-1)$.

After extracting the multi-scale time features using the common extractor and the first and last correlation features using the time-window transitional module, we splice the multiple features to perform the next step of electrode-related spatial feature extraction. The process of concat can be depicted in Eq. (1).

$$\boldsymbol{f}^{t}_{\mathrm{common}} = \begin{cases} \&concat\ \boldsymbol{f}^{t}_{depth}, \boldsymbol{f}_{start}, \boldsymbol{f}^{t}_{interval} & if\ t=1 \\ \&concat\ \boldsymbol{f}^{t}_{depth}, \boldsymbol{f}^{t-1}_{interval}, \boldsymbol{f}^{t}_{interval} & if\ 1<t\leq T-1 \\ \&concat\ \boldsymbol{f}^{t}_{depth}, \boldsymbol{f}^{t-1}_{interval}, \boldsymbol{f}_{end} & if\ t=T \end{cases} \quad (1)$$

where $\boldsymbol{f}^{t}_{\mathrm{common}} \in \mathbb{R}^{B\times V\times FE}(t = 1,2,\ldots T)$ is the output of each time window.

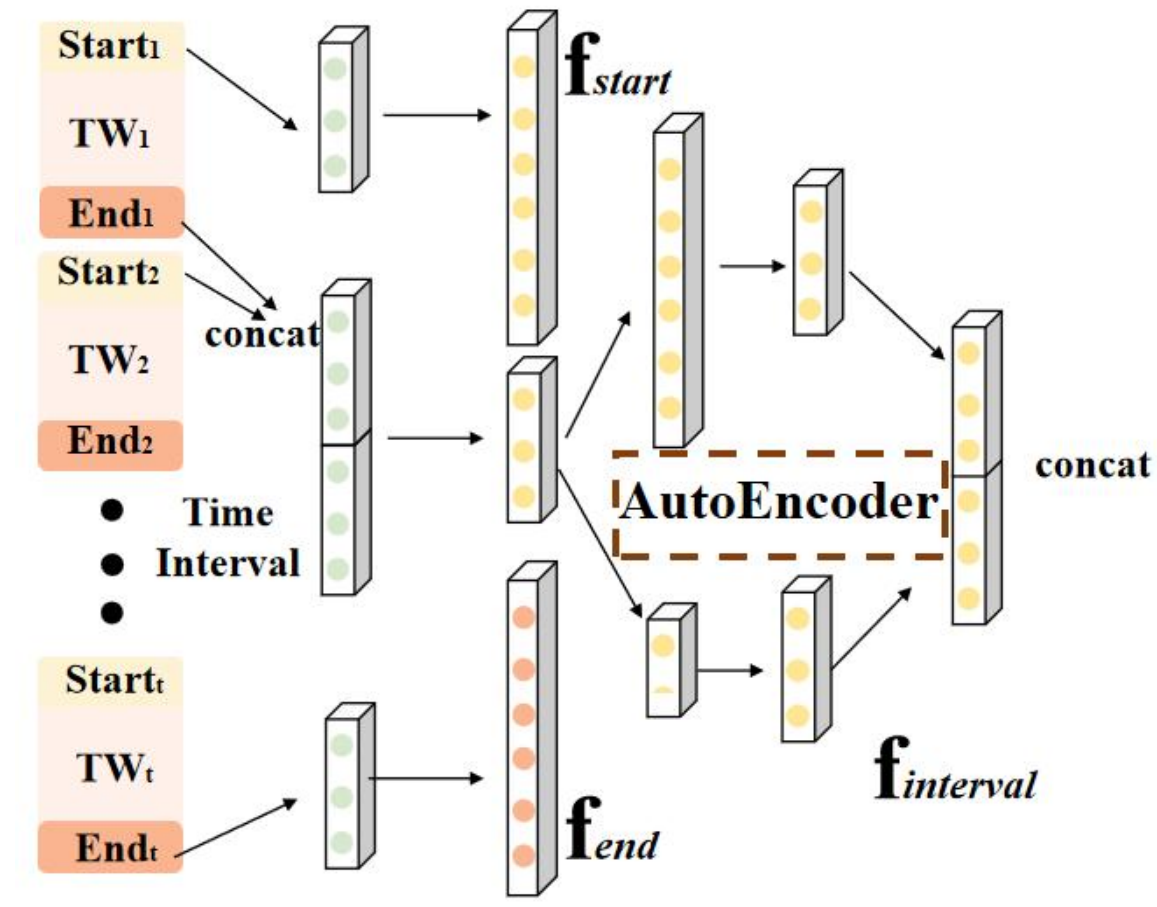

Figure 2. An overview of the time-window transitional module.

### c) *Electrode Spatial Topology Module*

Previous studies have shown that the connectivity patterns and activity intensity between different regions of the brain in patients with MDD undergo significant changes. For example, [17] indicates that MDD patients often exhibit functional connectivity abnormalities between the prefrontal cortex region (responsible for emotion control) and the limbic system (such as the amygdala, responsible for emotion production), which is considered as one of the neural foundations of emotion regulation disorders. Therefore, automatic weighting to the connection map by means of model training allows the model to learn enhanced spatial location information. In this part, we propose the electrode spatial topology module, which enables the model to automatically enhance the physical spatial connectivity by combining the electrode position information with a mask matrix. The physical spatial connectivity is fused with the functional connectivity. Then a GCN is used to aggregate innate spatial features based on the fused connectivity map.

In the electrode spatial topology module, we integrate EEG functional connectivity matrices with electrode physical adjacency matrix to embed spatial topological information. The physical adjacency matrix $\boldsymbol{A}_{Db} \in \mathbb{R}^{B\times V\times V}$ containing the spatial-position information of electrodes is used, as shown in Fig. 3. The connected state of $\boldsymbol{A}_{Db}$ is either connected or disconnected, which is respectively represented by 0 or 1. The λ-mask matrix $\boldsymbol{A}_{\lambda} \in \mathbb{R}^{B\times V\times V}$ is obtained by substituting the elements in $\boldsymbol{A}_{Db}$ whose values are 1 with $1+\lambda$ and 0 with $1-\lambda$. The matrix $\boldsymbol{A}_{\lambda}$ is introduced to fuse the $\boldsymbol{A}_{Db}$ with the functional connectivity map.

The functional connectivity map $\boldsymbol{A}_{FC} \in \mathbb{R}^{B\times T\times V\times V}$ is utilized to aggregate nodes' features inspired by graph attention networks [18], which can be represented in Eq. (2).

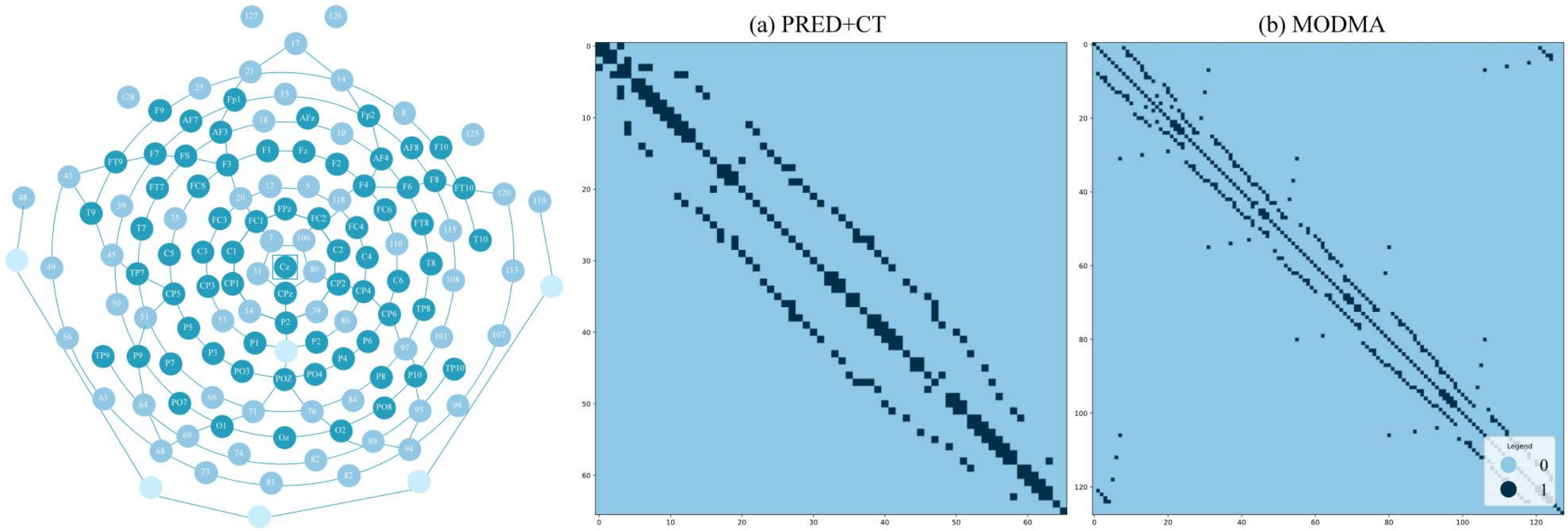

Figure 3. Schematic of the $\mathbf{A}_{Db}$ used to generate the $\lambda$-mask matrix. Where the left figure is the electrode physical connection diagram, and (a) and (b) on the right figure are the $\mathbf{A}_{Db}$ of two datasets.

$$\boldsymbol{A}^{t,h}_{F_{ij}} = \frac{exp\left(LeakyRelu\left(\left|\Phi^h(f^t_{v_i}) - \Phi^h\left(f^t_{v_j}\right)\right| \boldsymbol{a}\right)\right)}{\sum_{j=1}^{V} exp\left(LeakyRelu\left(\left|\Phi^h(f^t_{v_i}) - \Phi^h\left(f^t_{v_j}\right)\right| \boldsymbol{a}\right)\right)} \quad (2)$$

where $\boldsymbol{A}^{t,h}_{F_{ij}} \in \mathbb{R}^{B\times 1}$ means the value of the functional connectivity between two nodes $i$ and $j$ at the $t$-th time window under the $h$-th head, and $\boldsymbol{f}^t_{v_i}$ and $\boldsymbol{f}^t_{v_j} \in \mathbb{R}^{B\times FE}$ denote the common features of node $i$ and $j$ at the $t$-th time window, respectively. $\boldsymbol{a} \in \mathbb{R}^{FE\times 1}$ is a shared parameter in the multi-head attention mechanism, $\Phi^h(\cdot)$ stands for the $h$-th linear layer $(h = 1, 2, \ldots, H)$, whose projection does not change the last dimension of common features. $H$ denotes the head number.

The functional connectivity map at multiple time windows $\boldsymbol{A}_{FC}$ is derived from incorporating the connectivity maps at different heads, as shown in Eq. (3).

$$\boldsymbol{A}_{FC} = mean_{h=1,2,..,H}\boldsymbol{A}^h_F \quad (3)$$

The adjacency matrix $\boldsymbol{A}$ is obtained by fusing $\boldsymbol{A}_\lambda$ and $\boldsymbol{A}_{FC}$ using a Hadamard production $\otimes$ at each time window in Eq. (4).

$$\boldsymbol{A}^t = \boldsymbol{A}_\lambda \otimes \boldsymbol{A}^t_{FC}, \ \boldsymbol{A} = \underset{t=1,2,\ldots,T}{concat}\boldsymbol{A}^t \quad (4)$$

Then, we utilize the GCN to extract the spatial feature of $\boldsymbol{A}$ and $\boldsymbol{f}_{common}$ (This is the result of concatenating $\boldsymbol{f}^t_{\text{common}}$.) complexity in this process, which is shown in Eq. (5).

$$f_{est} = GCN(\boldsymbol{f}_{common}) \quad (5)$$

The output of the electrode distance connection module $\boldsymbol{f}_{est} \in \mathbb{R}^{B\times T\times V\times FS}$ is obtained after the GCN. It contains further innate spatial features under the fused connectivity. It is also a domain-invariant feature when applying the domain adaption method in this study.

### B. Secondary Time-Correlation Feature Extractor

To further mine the inherent temporal features from the extracted spatial features $\boldsymbol{f}_{est}$ and connect those time windows after electrode spatial topology module processing along the time window axis, we propose the secondary time correlation extraction module. This approach provides temporal features with high resolution and semantic information. Meanwhile, the long-range temporal dependencies of multiple time windows can be further exploited. The approach consists of two parts: the LSTM network and the graph transformer network (GTN) [19].

The LSTM network is used for multiple data windows containing spatial features, which in turn learns the complex long-range dependencies among them and identifies the periodic features about the MDD. The process is shown in Eq. (6).

$$\boldsymbol{f}_{LL} = ReLu\big(\Psi(\boldsymbol{f}_{est})\big) \quad (6)$$

where $\Psi(\cdot)$ is the sub-network function of LSTM.

We also use GTN network to extract temporal correlations, which breaks through the limitations of LSTM in extracting fixed temporal pattern features by dynamically constructing the graph structure and capture a longer node connection along the time-window dimension, explicitly representing the potential complex dependencies (edges) between time points. Different from the original GTN, which softly selects just one graph at each layer, we exploit weighted average for all the graphs after the Softmax operation. The process will enrich the temporal information of each time window, and assist with improving the effectiveness and accuracy of revealing temporal dependencies.

$$\boldsymbol{f}_{GTN} = GTN(\boldsymbol{f}_{est}) \quad (7)$$

The final feature map $\boldsymbol{f}_{Ste}$, containing innate temporal information after fusing time-windows, is derived by concatenating $\boldsymbol{f}_{LL}$ an $\boldsymbol{f}_{GTN}$ along the last feature dimension.

$$\boldsymbol{f}_{Ste} = \boldsymbol{f}_{LL} + \boldsymbol{f}_{GTN} \quad (8)$$

After that, $\boldsymbol{f}_{Ste}$ is put into a fully connected layer and a Softmax layer to obtain the predicted class labels. We utilize the cross-entropy function as the classifier loss, as shown in Eq. (9).

$$Loss_c = -\frac{1}{S}\sum_{i=1}^{S}\sum_{g=1}^{G} y_{i,g}\log \hat{y}_{i,g} \quad (9)$$

where $S$ denotes the number of subjects in the source domain while training in one batch, $G$ represents the number of classes and $y_{i,g}$ and $\hat{y}_{i,g}$ are the true label and the predicted class label (MDD, heathy control (HC)) of the subjects, respectively

### C. Domain Adaptation Module

We introduce domain adversarial learner in the DA module to enhance cross-subject MDD detection ability. The domain adversarial learner is composed of the domain adversarial neural network (DANN) [20] and assists in obtaining the domain-invariant features $\boldsymbol{f}_{est}$. The advantage of using gradient reversal layer, is that during the training of the network, the direction of the gradient passing through the gradient inversion layer is reversed, forcing the feature extraction layer of the network to learn features that are insensitive to changes in the domain (domain invariant feature), i.e., the model learns features that are more concerned with whether or not they are depressive, rather than focusing on individual subject feature. The gradient reversal layer loss function is shown in Eq. (10).

$$Loss_d = -\frac{1}{L\times T}\sum_{i=1}^{L}\sum_{t=1}^{T}\sum_{r=1}^{R}\mathcal{D}_{i,t,r}\log\widehat{\mathcal{D}}_{i,t,r} \tag{10}$$

where $L$ denotes the number of subjects in the source domain and target domain while training in one batch, $T$ indicates that $T$ time-windows are separately given the predicted labels, $R$ represents the number of classes and $\mathcal{D}_{i,t,r}$ and $\widehat{\mathcal{D}}_{i,t,r}$ are the true domain-label and the predicted domain-label (source domain or target domain), respectively.

In the training and updating process, both the MDD classifier and domain classifier want to obtain an accurate classification result. However, when using the gradient reversal layer, the gradient at the input of domain adversarial learner is reversed while backpropagation and the whole loss function for the network is equivalent to Eq. (11).

$$Loss = Loss_c - Loss_d \tag{11}$$

## III. EXPERIMENT

### A. Data Division

We used two publicly available MDD datasets (PRED+CT [21], MODMA [22]) to evaluate the effectiveness of our algorithm. For the experiments in this paper, a 10-fold cross-validation in cross-subject mode was adopted, and the data for the two datasets were divided in the following way. Firstly, for both datasets, the input of the model was obtained by cutting the original EEG data into $T$ non-overlapping time-windows. Secondly, when using 10-fold cross-validation in the cross-subject mode, the subjects were randomly divided into ten groups, and the number of subjects was even. However, considering that 52 subjects and 53 subjects were included in the PRED+CT dataset and MODMA dataset, respectively, two (2 = 52 mod 10) groups of the PRED+CT dataset and three (3 = 53 mod 10) groups of the MODMA dataset were randomly chosen to include one more subject in the tenfold cross-validation. The data of one group were treated as the test-set (target-domain) while the data of the other groups were treated as the train-set (source-domain).

### B. Platform and Evaluation Criterion

Table II. Hyperparameters of the proposed model.

| **Hyperparameters** | **Value (PRED+CT/MODMA)** |
|---|---|
| Batch size | 2 |
| Learning rate | 0.0001 |
| Epoch | 50 |
| Number of time-windows $T$ | 20 |
| Number of EEG channels | 66/128 |
| The length of each start/end slice | 125/75 sample points |
| Number of GCN kernels | 3 |
| Number of GT layers | 2 |

The proposed SET-TIME model was run on PyTorch 1.11.0 based on Python 3.8 under the NVIDIA GeForce GTX 3090 GPU. Furthermore, the SGD optimizer was used to train the model, and the hyperparameters of the model are shown in Table II. Apart from the commonly used criteria such as Accuracy and F1-score, the polygon area metric (PAM) [35] was introduced as an auxiliary metric, where a higher value also indicates a better performance like the former criteria.

### C. Comparisons with the Other Models

The experimental results in Table III show that the proposed SET-TIME outperforms the SOTA method on several metrics. On the PRED+CT and MODMA datasets, our method exceeds the SOTA method by 1.62% and 1.13% on the accuracy metric, respectively. The above experimental results show that our proposed SET-TIME model can effectively enhance the feature extraction capability by simultaneously considering the spatial information of electrode positions and the long-range dependency of time windows. To demonstrate the effectiveness of our proposed secondary temporal correlation feature extractor, we embedded it into a common feature extractor (Inter-STE) and found that the classification performance was significantly reduced. The reason may be that we performed secondary feature extraction on multiple time windows, which reduced the feature resolution and was not conducive to the feature alignment operation of the DA module.

TABLE III. Results of the comparison with the SOTA model. Best results are colored: first, second, third.

| **Dataset** | **Model** | **Accuracy (%)** | **F1-score (%)** | **PAM (%)** |
|---|---|---|---|---|
| PRED +CT | AMGCN-L [29] | 90.38 | 91.32 | 78.53 |
| | SSPA-GCN [30] | 83.17 | 82.93 | 65.69 |
| | GICN* [33] | 69.23 | 67.50 | 38.69 |
| | LSDD_EEGNet* [34] | 30.77 | 23.53 | 14.03 |
| | Inter-STE | 86.54 | 85.64 | 67.61 |
| | **SET-TIME** | **92.00** | **91.32** | **78.53** |
| MOD MA | GraphD [25] | 88.20 | 87.10 | \ |
| | DLMFD [26] | 92.13 | \ | \ |
| | AMG [27] | 88.68 | 88.17 | 76.65 |
| | DAN [28] | 87.40 | \ | \ |
| | DCANN [28] | 86.85 | 85.97 | \ |
| | AFL-CA [13] | 90.56 | 91.80 | 80.08 |
| | GCN-Transformer | 89.03 | 88.83 | \ |
| | AMGCN-L [29] | 90.57 | 90.51 | 81.03 |
| | SSPA-GCN [30] | 92.87 | 92.12 | 83.63 |
| | GICN* [33] | 67.92 | 62.60 | 21.96 |
| | LSDD_EEGNet* [34] | 53.85 | 35.00 | 42.41 |
| | Inter-STE | 75.38 | 72.80 | 46.46 |
| | **SET-TIME** | **94.00** | **93.94** | **85.76** |

Where * represent the reproduced results, and the Inter-STE is the model that the secondary time-correlation feature extractor in the common feature exactor.

### D. Ablation Study

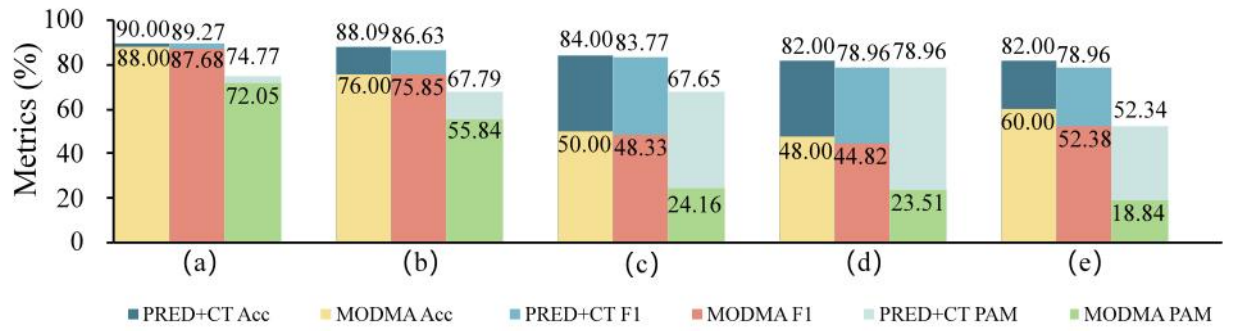

Figure 4. Results of the experiments on ablation study.

A set of ablation studies was conducted to investigate the effectiveness of the components in the proposed SET-TIME model. The following components were deleted in turn while in the ablation studies: (a) without time-window transitional module, (b) without electrode spatial topology module, (c) without secondary time-correlation feature extractor, (d) deleting the LSTM in secondary time-correlation feature extractor, (e) deleting the GTN in secondary time-correlation feature extractor.

After removing the time-window transitional module from SET-TIME, the results of both datasets decreased, proving the effectiveness of the time-window transitional module, which not only extracts the long-range dependencies between windows, but also captures the mutation signal changes within windows. When the electrode spatial topology module was deleted, a more significant performance drop was observed on both datasets compared to the experiment where the time window correlation was removed. Particularly for the MODMA dataset, which utilizes 128 EEG electrode channels, the Acc dropped to just 76.00%, and the PAM to 55.84%. These reductions represent a decrease of 18.00% and 29.92%, respectively, compared to the model's peak performance. This phenomenon indicates that the electrode topology connection plays a crucial role in processing common features in the electrode topology connection module by fusing different views of connectivity maps and suggests that its impact may be related to the number of EEG signal channels. The substantial influence of the electrode topology connection removal, especially on the MODMA dataset, underscores its importance in effectively handling the complexity and diversity of features derived from a larger number of EEG channels, thereby affirming its significance in enhancing model performance through fusion of the different views of the spatial connectivity maps.

The results regarding the removal of the secondary time-correlation feature extractor present a more complex and unexpected scenario. Initially, it was hypothesized that the performance impact of removing either the LSTM or the GTN component would be significantly less detrimental than removing both, essentially eliminating the entire secondary time correlation extraction module. However, findings reveal that for the PRED+CT dataset, the outcomes of removing the LSTM were equivalent to those of removing the GTN, and though close to the results of eliminating the secondary time-correlation feature extractor entirely, the removal of the entire secondary time correlation extraction module surprisingly yielded slightly better results.

For the MODMA dataset, from the perspective of PAMs (with PAM for ablation study (c), (d), and (e) being 24.16%, 23.51%, and 18.84%, respectively), the best performance was unexpectedly observed by removing both components. Yet, when considering accuracy and F1-score, removing the LSTM did not perform as well as removing the entire secondary time correlation extraction module, while removing the GTN resulted in better outcomes than eliminating the secondary time correlation extraction module. The only result consistent with initial expectations, where removing GTN leads to better performance than removing the secondary time correlation extraction module, was observed in the MODMA dataset from this specific perspective. In the PRED+CT dataset, regardless of which performance metric is considered, the effect of removing the entire secondary time-correlation feature extractor was counterintuitively found to be more beneficial than removing just one of its parts. This phenomenon suggests that the interconnection between these temporal windows is intricate and that the combination of LSTM with the modified GTN is essential to fuse those time-windows successfully and achieve good performance. It also highlights the complex nature of temporal information fusion in EEG signal analysis and suggests that the synergistic effect of combining these two components is critical for capturing the nuanced temporal dynamics within the data.

### E. Visualization Analysis

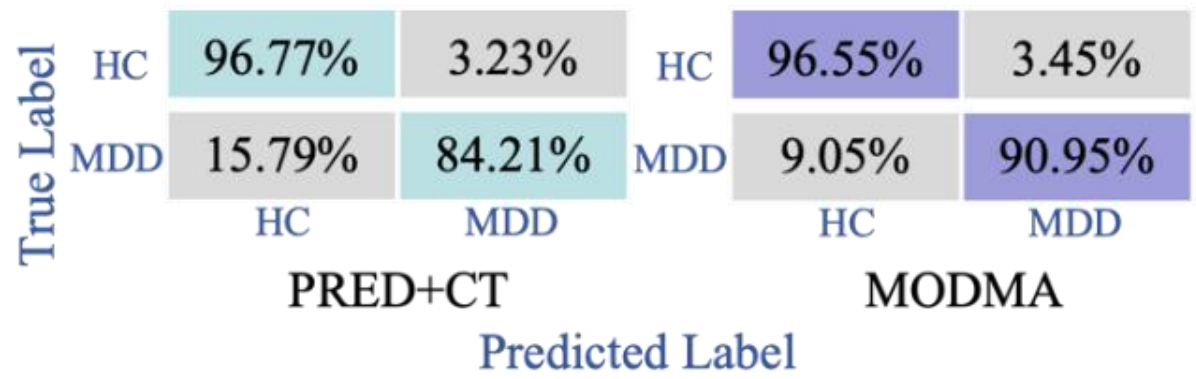

Figure 5. Confusion Matrices on PRED+CT and MODMA Datasets.

Following the indication of existing studies [23] [31] [32], we visualized the effectiveness of our approach using a confusion matrix. As shown in Fig. 5, we quantitatively evaluated the recognition results by analyzing the classification effect of the SET-TIME model on the two datasets with different categories through the confusion matrix. The classification effect of HC was 96.77% on the PRED+CT dataset and 96.55% on the MODMA dataset. However, in the MDD category, the classification effect of the PRED+CT dataset was not as good as that of the MODMA dataset, which may be caused by the fact that, the PRED+CT dataset used fewer number of electrode channels and lacked spatial connectivity information, which resulted in the model not being able to capture more features about the MDD disease-related brain regions. This further proves the effectiveness of the method proposed in this paper and its positive effect on model performance.

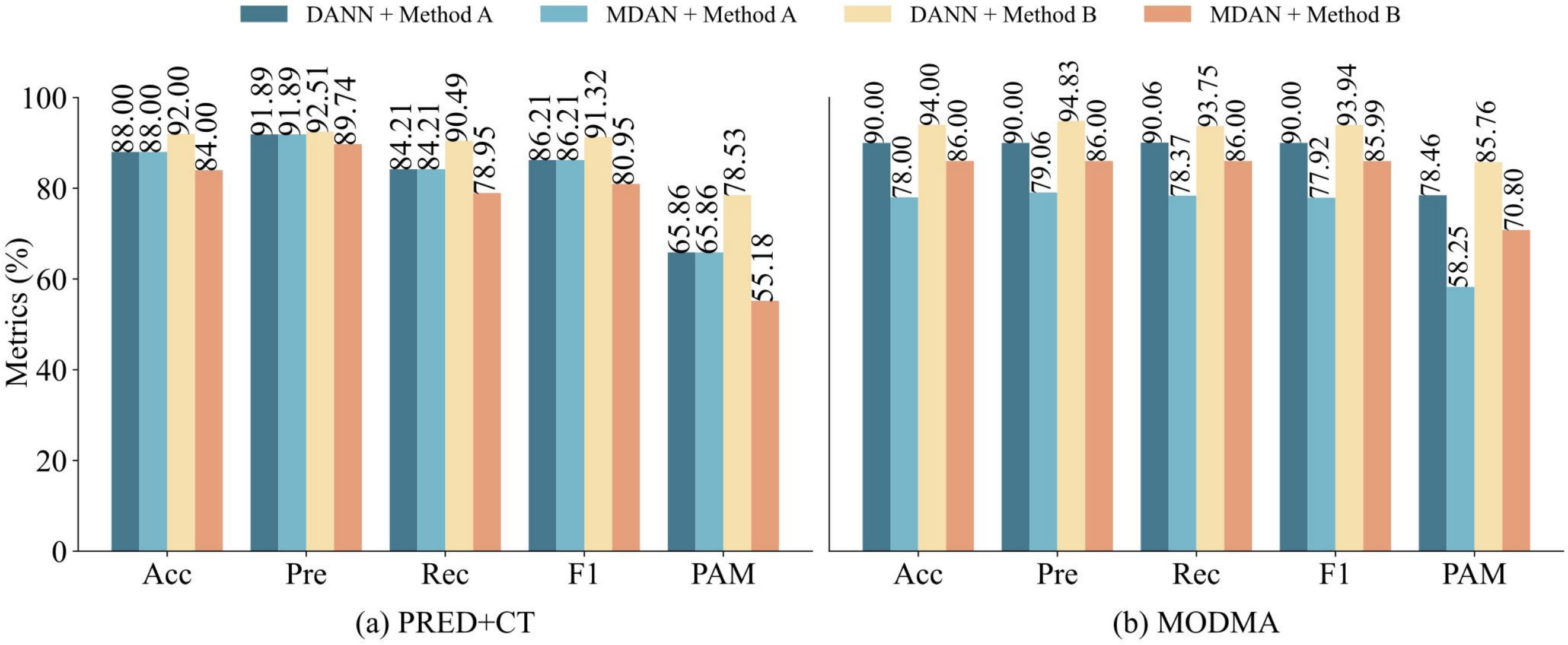

Figure 6. Results of the experiments on domain-related structure obtained from the PRED+CT and the MODMA datasets.

## IV. Discussion

### A. Analysis of the Importance of Domain Invariant Features in Different Areas

To verify the importance of domain invariant feature, we choose the output of common feature extractor (Method A) and the output of the proposed electrode spatial topology module (Method B) as the domain invariant features. Here, domain-invariant features refer to those that are fed into the domain adversarial learner to align feature distributions between source and target domains. Furthermore, we also introduce in the DA method: Multi-Domain Adversarial Network (MDAN) [24], considering multiple time windows have been used. The experimental results are shown in Fig. 6. On the MODMA dataset, the classification accuracies using the DANN + Method B and MDAN + Method B combination are higher than the DANN + Method A and MDAN + Method A combination. However, on the PRED+CT dataset, the MDAN + Method B combination showed a decrease in classification accuracy compared to MDAN + Method A. This may be attributed to the small number of electrodes in the PRED+CT dataset, resulting in insufficient extracted electrode spatial topology to adequately characterize the valid domain invariant information. In addition, when using the MDAN method and letting its multiple domain classifiers deal with different time windows separately, due to the specificity of each time window itself, this multi-classifier counter-training mechanism may instead interfere with the learning process of the model, which ultimately leads to a degradation of the model performance.

For both PRED+CT and MODMA datasets, the best results were achieved by using DANN + Method B, and robust classification accuracy was realized. Although the results of MODMA dataset do not necessarily exceed those of PRED+CT dataset when MDAN is used, and the results of MODMA dataset are usually better than those of PRED+CT dataset when DANN is applied, the best performance of the two datasets is still similar, where the best result of PRED+CT is 92.00% accuracy, which is close to the best result of MODMA dataset of 94.00% accuracy, indicating the stability of SET-TIME.

### B. Multiple Time Window Feature Visualization and Analysis

To further explore the specificity of time windows, we visualized 20 time windows using t-SNE, including t-SNE feature maps of one MDD subject and one HC subject randomly selected from the same target domain. The experimental results proved our conjecture. As shown in Fig. 7 (a), for the PRED+CT dataset, time windows 3, 5, 6, 9, 15, and 19 still maintain a relatively clear demarcation, but time windows 2, 11, 14, 16, and 20 show significant feature intermingling, while the other time windows also show more or less feature intermingling. As shown in Fig. 7 (b), for the MODMA dataset, although there are two piles of data with obvious demarcation in each time window, the feature points of the other piles of MDD data appear clearly in one of the HC piles, i.e., the phenomenon of feature fusion occurs. The appearance of this phenomenon is precisely due to DANN's role: its domain classifier guides the direction of feature updates, leading to the alignment of target domain features with source domain features, so there will be the above feature fusion.

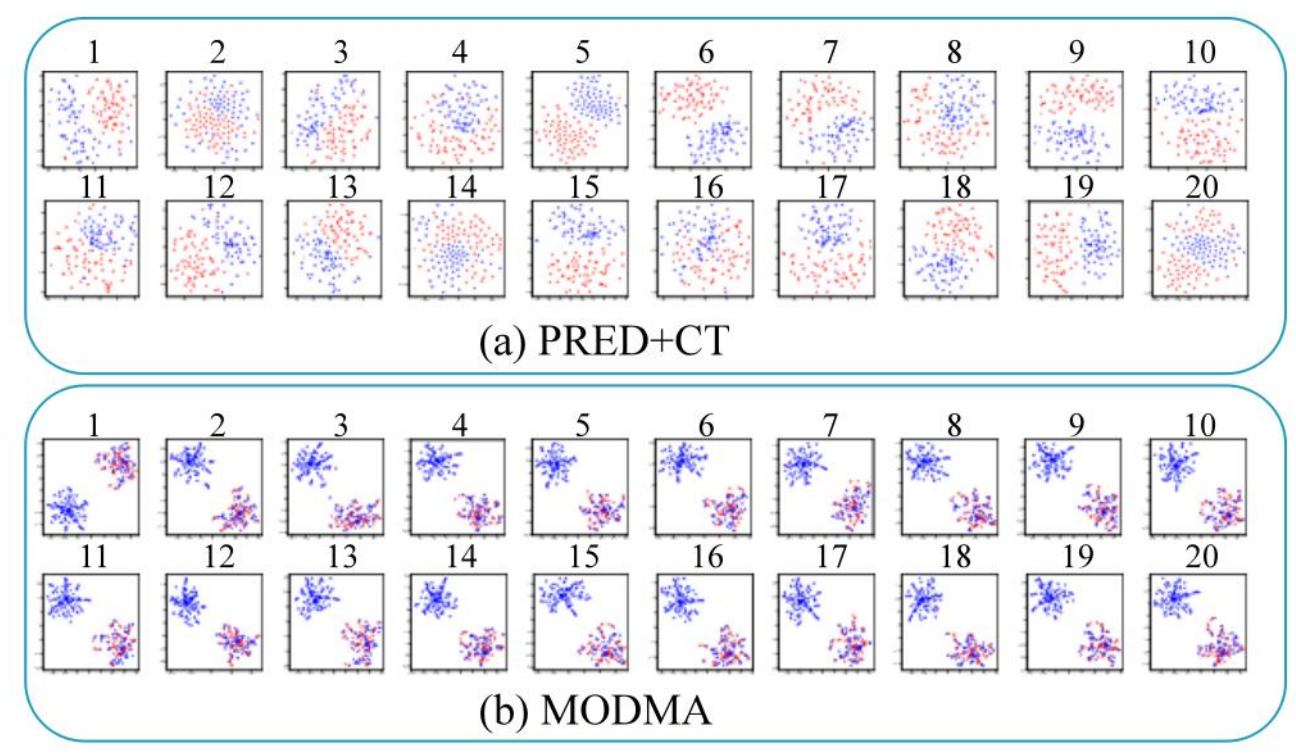

Figure 7. t-SNE feature maps of two time-windows of spatial feature (Red: HC, Blue: MDD).

### C. Secondary Time-Correlation Feature Extractor Visualization and Analysis

Fig. 8 shows that after extracting the temporal information from multiple time windows using the secondary time-correlation feature extractor, a clear demarcation between subjects in the two datasets is observed in the t-SNE map, especially in the MODMA dataset. This observation suggests that the secondary time-correlation feature extractor successfully fuses different time windows and extracts deep temporal information from the EEG signals, which enhances the discriminative rows of the features and is helpful for the final classification results.

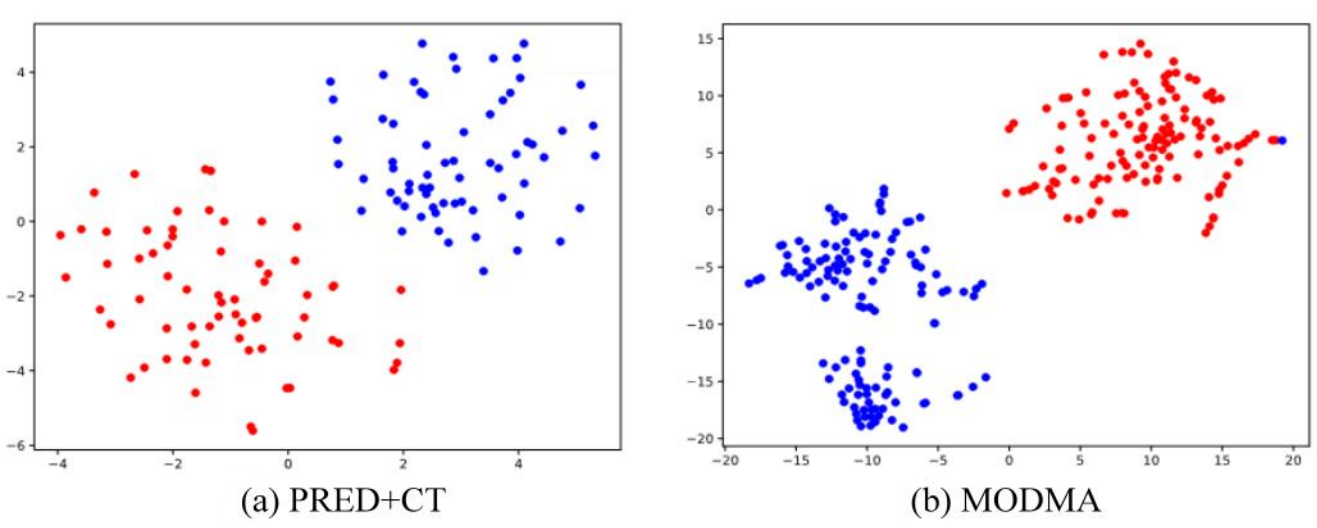

(a) PRED+CT (b) MODMA

Figure 8. t-SNE feature map after secondary time-correlation feature extracted (Red: HC, Blue: MDD).

## V. CONCLUSION

In this work, a so-called SET-TIME deep learning model that considers the electrode spatial topology and time-window transition of EEG signals is put forward for MDD detection. SET-TIME comprises a novel common feature extractor with a multi-scale receptive field and a time window transitional module, an electrode spatial topology module, and a secondary time-correlation feature extractor. We also embed a DA module into our proposed SET-TIME network to improve the cross-subject detection performance. Extensive experiments have been conducted in the individual-based tenfold cross-validation for depression detection with the cross-subject mode. SET-TIME has achieved an accuracy of 92.00% and 94.00% on PRED+CT and MODMA datasets for the MDD detection task, respectively. From these experimental results, it can be seen that a common feature extractor that preserves multi-scale information of channels, physical connectivity relationships between multiple EEG channels, and correlation information of multiple time windows can generate powerful features for MDD detection. Moreover, it has proved from the results that the combination of the secondary time correlation extraction would help to fuse the information from multiple time windows of the depressed and healthy subjects properly.

## ACKNOWLEDGMENT

This research is supported by the National Natural Science Foundation of China (Grant Nos 62203321 and 62571368).